\documentclass[10pt,twocolumn,letterpaper]{article}

\usepackage{wacv}
\usepackage{times}
\usepackage{epsfig}
\usepackage{graphicx}
\usepackage{amsmath}
\usepackage{amssymb}
\usepackage{booktabs}
\usepackage{adjustbox}
\usepackage{caption}
\usepackage[accsupp]{axessibility}
\usepackage{tabularx, booktabs}
\newcolumntype{?}{!{\vrule width 1.2pt}}
\newcolumntype{Y}{>{\centering\arraybackslash}X}

\def\wacvPaperID{615} 

\wacvalgorithmstrack   

\wacvfinalcopy 

\ifwacvfinal
\usepackage[breaklinks=true,bookmarks=false]{hyperref}
\else
\usepackage[pagebackref=true,breaklinks=true,colorlinks,bookmarks=false]{hyperref}
\fi

\begin{document}

\title{FaceOff: A Video-to-Video Face Swapping System}

\author{Aditya Agarwal$^*$\\
IIIT Hyderabad\\
{\tt\small aditya.ag@research.iiit.ac.in}

\and
Bipasha Sen$^*$\\
IIIT Hyderabad\\
{\tt\small bipasha.sen@research.iiit.ac.in}

\and
Rudrabha Mukhopadhyay\\
IIIT Hyderabad\\
{\tt\small radrabha@research.iiit.ac.in}

\and
Vinay Namboodiri\\
University of Bath\\
{\tt\small vpn22@bath.ac.uk}

\and
C V Jawahar\\
IIIT Hyderabad\\
{\tt\small jawahar@iiit.ac.in}
}

\twocolumn[{
\maketitle
\begin{center}
\includegraphics[width=\textwidth]{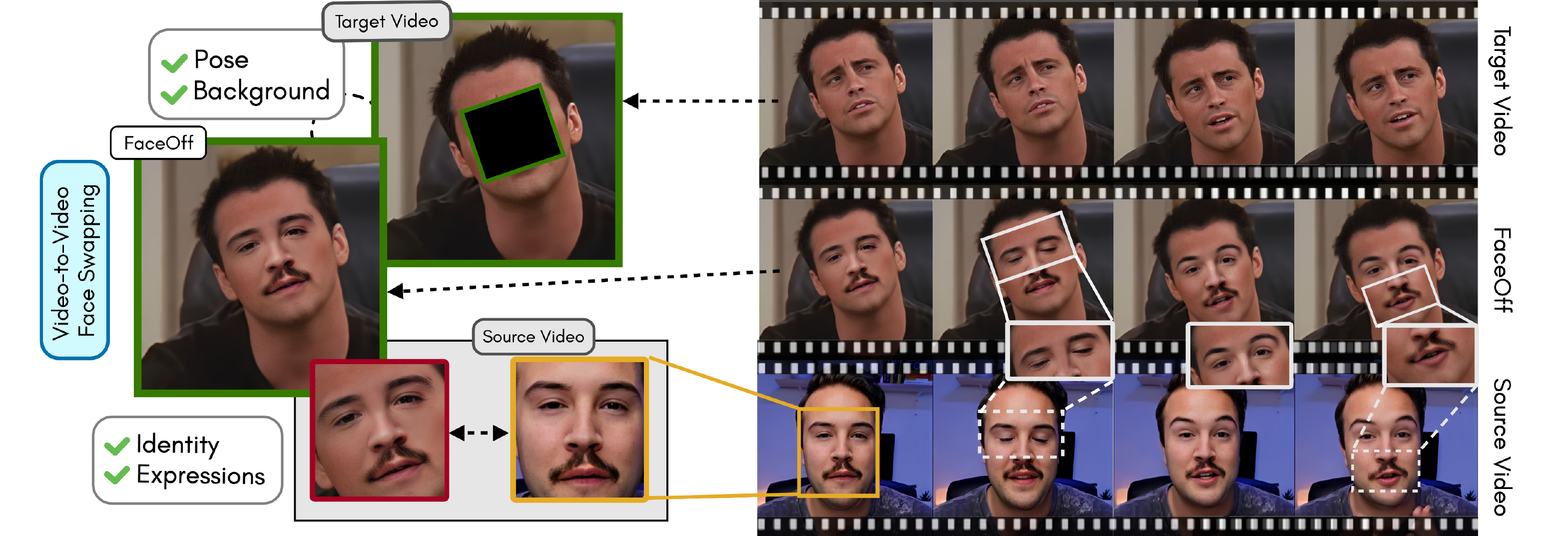}
\captionof{figure}{\small We introduce video-to-video (V2V) face-swapping, a novel task of face-swapping that aims to swap the \textbf{identity} and \textbf{expressions} from a source face \underline{video} to a target face \underline{video}. This differs from the face-swapping task that aims to swap only an identity. There are many downstream applications of V2V face-swapping, such as automating the process of an actor replacing their double in movie scenes, which today is handled manually using expensive CGI technology. In this example, Nolan, an actor (source video), is recording his dialogues and expressions at the convenience of his home. Joey Tribiani (target video) is acting as his double in a scene of the famous sitcom FRIENDS. FaceOff face-swaps Nolan into the scene. Please note the zoomed-in source (yellow box) and face-swapped (red box) output. In this output, although the source face pose and skin complexion have changed and blended with the background, identity and expressions are preserved.}

\label{fig:teaser}
\end{center}
}]
\def\thefootnote{*}\footnotetext{Equal contribution}
\thispagestyle{empty}

\begin{abstract}

Doubles play an indispensable role in the movie industry. They take the place of the actors in dangerous stunt scenes or scenes where the same actor plays multiple characters. The double's face is later replaced with the actor's face and expressions manually using expensive CGI technology, costing millions of dollars and taking months to complete. An automated, inexpensive, and fast way can be to use face-swapping techniques that aim to swap an identity from a source face video (or an image) to a target face video. However, such methods cannot preserve the source expressions of the actor important for the scene's context. 
To tackle this challenge, we introduce video-to-video (V2V) face-swapping, a novel task of face-swapping that can preserve (1) the identity and expressions of the source (actor) face video and (2) the background and pose of the target (double) video. 
We propose FaceOff, a V2V face-swapping system that operates by learning a robust blending operation to merge two face videos following the constraints above. It reduces the videos to a quantized latent space and then blends them in the reduced space. FaceOff is trained in a self-supervised manner and robustly tackles the non-trivial challenges of V2V face-swapping. As shown in the experimental section, FaceOff significantly outperforms alternate approaches qualitatively and quantitatively.

\end{abstract}

\section{Introduction}

Having doubles\def\thefootnote{$^1$}\footnote{\href{https://en.wikipedia.org/wiki/Double\_(filmmaking)}{https://en.wikipedia.org/wiki/Double\_(filmmaking)}} for the starring actors in movies is an indispensable component of movie-making.
A double may take the actor's place during stunt scenes involving difficult and dangerous life-risking acts. They may even stand-in for the actor during regular fill scenes or multiple retakes.
For instance, `The Social Network' extensively used body doubles as a stand-in for actor Armie Hammer who played multiple roles of twin brothers\def\thefootnote{$^2$}\footnote{\href{https://www.youtube.com/watch?v=spIdefyvjTs}{Captain America - Skinny Steve Rogers Behind the Scenes}}\def\thefootnote{$^3$}\footnote{\href{https://www.youtube.com/watch?v=fCrYfRjpuXU&t=26s}{How CGI made Cody and Caleb as PAUL WALKER | VFX}}\def\thefootnote{$^4$}\footnote{\href{https://www.cinemablend.com/new/Armie-Hammer-Didn-t-Play-Both-Winklevoss-Twins-Social-Network-20994.html}{Armie Hammer Didn't Play Both Winklevoss Twins Social Network}}.
In such scenes, the double's face is later replaced by the actor's face and expressions using CGI technology requiring hundreds of hours of manual multimedia edits on heavy graphical units costing millions of dollars and taking months to complete.
Thus, the production team is generally forced to avoid such scenes by changing the mechanics of the scene such that only the double's body is captured to provide an illusion of the actor. This may act as a constraint on the director's creativity. However, such adjustments are not always possible. 

A different scenario is post-production scene modifications. If a dialogue is discovered in post-production that suits a scene better than the original, the entire scene is reset and re-shot. 
We propose that the actor could instead record in a studio and get their face superimposed on the previous recording. 
In fact, like other industries, the movie industry is also headed in the direction where actors can work from home. In today's era, CGI technologies can produce incredible human structures, scenes, and realistic graphics. However, it is known that they struggle to create realistic-looking skin\def\thefootnote{$^5$}\footnote{\href{https://www.youtube.com/watch?v=FtifBqf2Z50}{Why It's SO HARD To Do CGI Skin!}}. As shown in Fig.~\ref{fig:teaser}, an actor could lend their identity and expressions from the comfort of their home or studio while leaving the heavy-duty to graphics or a double. 
CGI technologies needed for such tasks are manually operated, expensive, and time-consuming. 

To automate such tasks, fast and inexpensive computer vision based face-swapping~\cite{deepfacelabs, motion-coseg, fsgan, faceswapdisney, faceshifter, faceswapphotos} techniques that aim to swap an identity between a source (actor) video and target (double) video can be considered. However, such techniques cannot be directly used. Face-swapping swaps only the source identity while retaining the rest of the target video characteristics. In this case, the actor's expressions (source) are not captured in the output. 
To tackle this, we introduce "video-to-video (V2V) face-swapping" as a novel task of face-swapping that aims to \textbf{(1)} swap the identity and expressions of a source face video and \textbf{(2)} retain the pose and background of the target face video. 
The target pose is essential as it depends on the scene's context. 
E.g., a stunt man performs at an outdoor location dealing with machines or talking to a fellow double; the actor acts in front of a green screen at a studio. Here, the double's pose is context-aware, and the actor only improvises. 

\textbf{How is the proposed task a video-to-video face-swapping task? }Unlike the face-swapping task that swaps a fixed identity component from one video to another video, V2V face-swapping swaps expressions changing over time (a video) with another video with changing pose and background (another video), making our task video-to-video. 

\textbf{Approach: }Swapping faces across videos is non-trivial as it involves merging two different motions - the actor's face motion (such as eye, cheek, or lip movements) and the double's head motion (such as pose and jaw motion). This needs a network that can take two different motions as input and produce a third coherent motion.
We propose \textbf{FaceOff}, a video-to-video face swapping system that reduces the face videos to a quantized latent space and blends them in the reduced space. 
A fundamental challenge in training such a network is the absence of ground truth.
Face-swapping approaches~\cite{motion-coseg, fsgan, deepfacelabs} use a discriminator-generator setup for training the networks.
The discriminator is responsible for monitoring the desired characteristic of the swapped output. However, using a discriminator leads to hallucinating components of the output different from the input - for instance, modified identity or novel expressions.
Thus, we devise a self-supervised training strategy for training our network: We use a single video as the source and target. We then introduce pseudo motion errors on the source video. Finally, we train a network to `fix' these pseudo errors to regenerate the source video. 

FaceOff can face-swap unseen cross-identities directly at inference without any finetuning. 
Moreover, unlike most face-swapping methods that need inference time optimization ranging from $5$ minutes to $24$ hours on high-end GPUs, FaceOff face-swaps videos in just one forward pass, taking less than a second. A key feature of FaceOff is that it preserves at least one of the input expressions (source in our case), whereas, as we show later, existing methods fail to preserve either of the expressions (source or target expressions). 
Lastly, we curate and benchmark V2VFaceSwap, a V2V face-swapping test dataset made of instances from unconstrained YouTube videos on unseen identities, backgrounds, and lighting conditions.

\textbf{Our contributions} in this work are as follows: (1) We introduce V2V face-swapping, a novel task of face-swapping that aims to swap source face identity and expressions while retaining the target background and pose. (2) We propose FaceOff: a V2V face-swapping system trained in a self-supervised manner. FaceOff generates coherent videos by merging two different face videos. (3) Our approach works on unseen identities directly at the inference time without any finetuning. (4) Our approach does not need any inference time optimization taking less than a second to infer. (5) We release the V2VFaceSwap test dataset and establish a benchmark for the V2V face-swapping task.

\section {Related Work}

Table~\ref{tab:comps} provides a comparison between the existing tasks and FaceOff. FaceOff aims to solve a unique challenge of V2V face-swapping that has not been tackled before. 

\textbf{Face Swapping}: Swapping faces across images and videos have been well-studied \cite{deepfacelabs, fsgan, motion-coseg, simswap, fastfaceswap, faceshifter, faceswapdisney, faceswapphotos, 3dmodelfaceswapping} over the years. These works aim to swap an identity obtained from a source video (or an image) with a target video of a different identity such that all the other target characteristics are preserved in the swapped output.
DeepFakes\def\thefootnote{$^6$}\footnote{\href{https://github.com/deepfakes/faceswap}{https://github.com/deepfakes/faceswap}}, DeepFaceLabs~\cite{deepfacelabs}, and FSGAN~\cite{fsgan} swap the entire identity of the source; 
Motion-coseg~\cite{motion-coseg} specifically swaps the identity of single/multiple segments of a given source image (either hair or lips or nose, etc.) to a target video.
Unlike these approaches that swap only the identity or a specific part of an image, we swap temporally changing expressions along with the identity of the source. 
Moreover, FSGAN takes $5$ minutes of inference time optimization, DeepFaceLabs and DeepFakes take up to $24$ hours of inference time optimization on high-end GPUs. FaceOff takes less than a second to face swap in-the-wild videos of unseen identities.

\begingroup
\renewcommand{\arraystretch}{1.1} 
\begin{table}[t]
  \resizebox{\linewidth}{!}{
  \begin{tabular}{|l|c|c|c|c|}
  \hline
    & \multicolumn{2}{c}{\textbf{Source}} & \multicolumn{2}{|c|}{\textbf{Target}}\\
    \cline{2-5}
    \textbf{Method} & \textbf{Identity} & \textbf{Expression} & \textbf{Pose} & \textbf{Background} \\
    \hline
    Face Swapping 
    & \checkmark  & $\times$ & \checkmark & \checkmark \\
    \hline
    Face Reenactment 
    & $\times$ & \checkmark & $\times$ & \checkmark \\
    \hline
    Face Editing 
    & $\times$ & $\times$ & \checkmark & \checkmark \\
    \hline
    FaceOff (Ours) & \checkmark & \checkmark & \checkmark & \checkmark \\
    \hline
  \end{tabular}
  }
  \caption{Comparison of FaceOff with existing tasks. \checkmark and $\times$ indicate if the characteristic is preserved and lost respectively. FaceOff solves a unique task of preserving source identity and expressions that has not been tackled before.}
\label{tab:comps}
\end{table}
\endgroup

\textbf{Face Manipulation}: Face manipulation animates the pose and expressions of a target image/video according to a given prior~\cite{face-vid2vid, fomm2, fomm, reenactgan, deepfacelabs, flowguided, nvp, makeittalk}.
In audio-driven talking face generation~\cite{wav2lip, lipgan, wav2lip-emotion, posecont, nvp, pirenderer, vdub}, the expressions, pose, and lip-sync in the target video are conditioned on a given input speech audio. Unlike such works, we do not assume an audio prior for our approach. 
A different direction of \textbf{face reenactment} animates the source face movements according to the driving video \cite{deffererneuralrendering, pirenderer, face2face, deepvideopotraits, fomm, fomm2}.
The identity is not exchanged in these works. This can tackle a special case of our task -- when the target and source have the same identity. Here, a target image can be re-enacted according to the source video expressions. As we show in Section~\ref{sec:targetfacemanipulation}, FaceOff captures the micro-expression of the driving video, unlike the existing approaches.
This is because we rely on a blending mechanism - allowing a perfect transfer of the driving expressions.
Another direction that handles this special case is \textbf{face editing}, which involves editing the expressions of a face video. Using this approach, one can directly edit the target video according to the source expressions. 
Image-based face editing works such as \cite{pix2pix,stargan,starganv2,cgan} have gained considerable attention. 
However, realizing these edits on a sequence of frames without modeling the temporal dynamics often results in temporally incoherent videos. Recently, STIT~\cite{stit} was proposed that can coherently edit a given video to different expressions by applying careful edits in the video's latent space. 
Despite the success, these techniques allow limited control over expression types and variations. Moreover, obtaining a correct target expression that matches the source expressions is a manual hit and trial. FaceOff can add micro-expressions undefined in the label space simply by blending the emotion from a different video of the same identity with the desired expressions.

\section{FaceOff: Face Swapping in videos}
\begin{figure*}[t]
  \centering
  \includegraphics[width=1.0\linewidth]{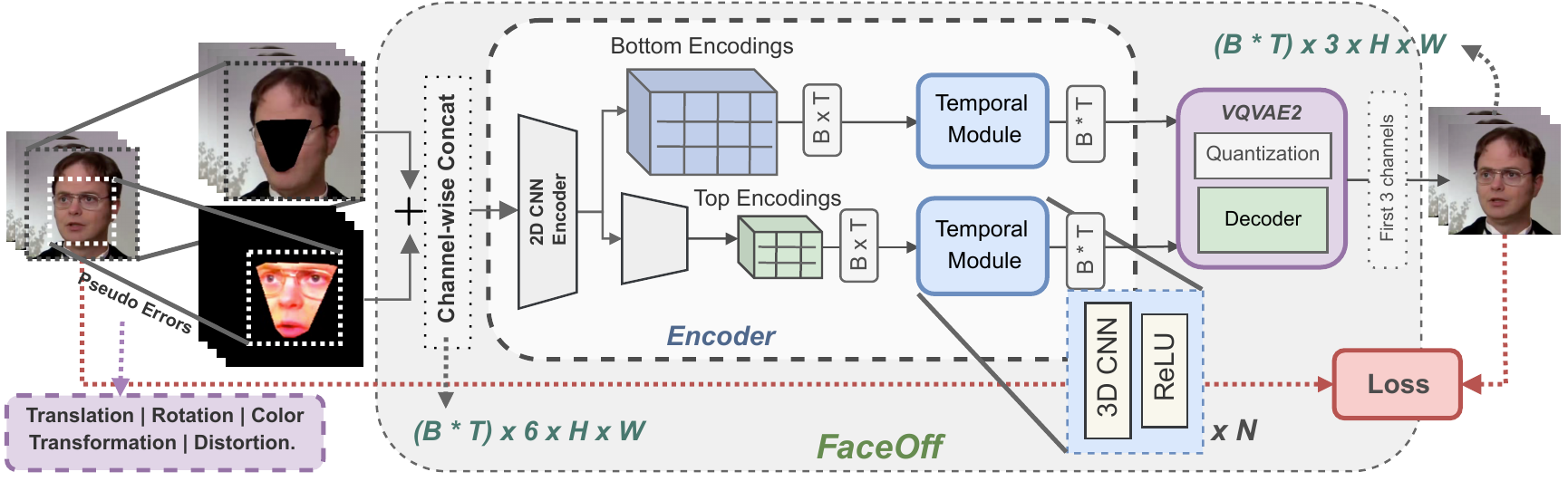}
  \caption{FaceOff is a temporal autoencoder operating in a hierarchical quantized latent space. We use a self-supervised training scheme to train FaceOff using a distance loss on the exact output-ground truth pairs. In the scheme, we first extract the face, $f$, and background, $b$, from a single video, $s$. We then apply ``pseudo errors" made of random rotation, translation, scaling, colors, and non-linear distortions to modify $f$. Next, modified $f$ (acting as a source) and $b$ (acting as a target) are concatenated at each corresponding frame channel-wise to form a single video input. This video input is then reduced and blended, generating a coherent and meaningful output. This output is expected to match the source video, $s$. }
  
  \label{fig:arch_main}
\end{figure*}
\begin{figure}[t]
  \centering
  \includegraphics[width=\linewidth]{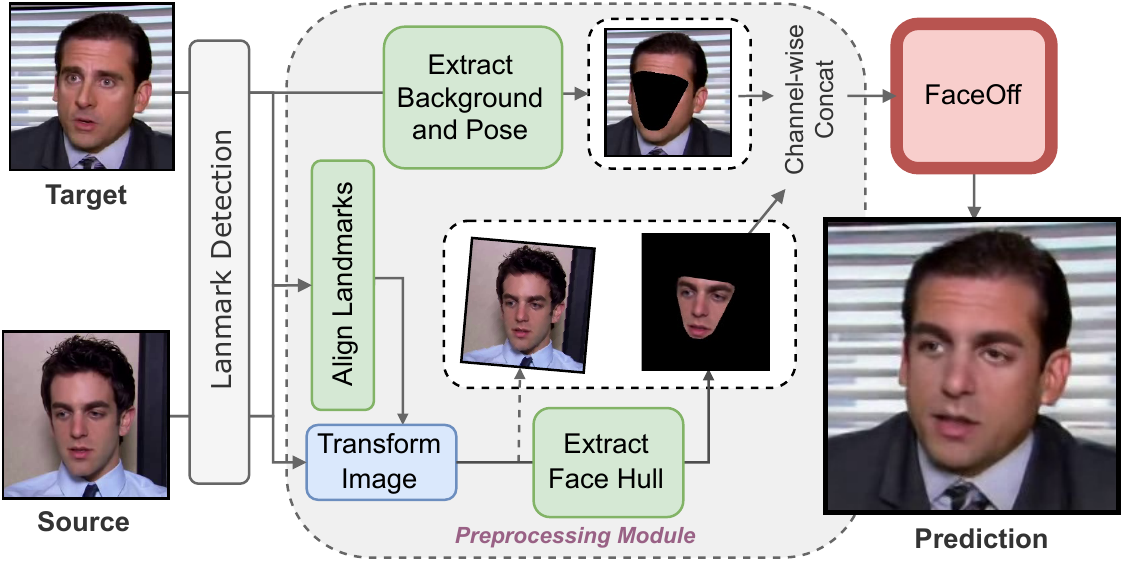}
  \caption{Inference pipeline: FaceOff can be directly inferred on any unseen identity without any finetuning.
  At inference, the source video is first aligned frame-by-frame using the target face landmarks. FaceOff then takes (1) foreground of the aligned source video and (2) background and pose of the target video as input and generates the output.}
  \label{fig:inference_pipe}
\end{figure}

We aim to swap a source face video with a target face video such that (1) the identity and the expression of the source video are preserved and (2) the pose and background of the target video are retained. To do this, we learn to blend the foreground of the source face video with the background and pose of the target face video (as shown in Fig.~\ref{fig:inference_pipe}) such that the blended output is coherent and meaningful. This is non-trivial as it involves merging two separate motions.
Please note that we only aim to blend the two motions; thus, the desired input characteristics -- identity, expressions, pose, and background -- are naturally retained from the inputs without additional supervision. The main challenge is to align the foreground and background videos so that the output forms a coherent identity and has a single coherent pose. All the other characteristics are reconstructed from the inputs. 
Our core idea is to use a temporal autoencoding model that merges these motions using a quantized latent space.
Overall, our approach relies on (1) Encoding the two input motions to a quantized latent space and learning a robust blending operation in the reduced space.
(2) A temporally and spatially coherent decoding. 
(3) In the absence of ground truth, a self-supervised training scheme.

\subsection{Merging Videos using Quantized Latents}
\label{sec:merging_video_using}
We pose face-swapping in videos as a blending problem: given two videos as input, blend the videos into a coherent and meaningful output. We rely on an encoder to encode the input videos to a meaningful latent space. Our overall network is a special autoencoder that can then learn to blend the reduced videos in the latent space robustly and generate a blended output. We select our encoder model carefully, focusing on ``blending" rather than learning an overall data distribution. Encoder networks with a continuous latent space reduce the dimension of a given input, often down to a single vector that can be considered a part of an underlying distribution. This latent vector is highly stochastic; a very different latent is generated for each new input, introducing high variations that a decoder needs to handle. Recently, ``vector quantization" was proposed in \cite{vqvae, vqgan, vqvae2}. Quantization reduces the variation in latents by fixing the number of possible latent codes. However, retaining the input properties using a single quantized latent vector is impossible. Thus the inputs are reduced to a higher dimensional quantized space (such as $64 \times 64$) such that properties of the input needed for a full reconstruction are preserved. We adopt such an encoder in our proposed autoencoder for encoding our videos. As shown in Fig.~\ref{fig:arch_main}, our encoder is a modified VQVAE2~\cite{vqvae2} encoder that encodes videos instead of images. We introduce temporal modules made of non-linear 3D convolution operations to do so. 

The input to our encoder is a single video made by concatenating the source foreground and target background frames channel-wise, as shown in Fig.~\ref{fig:inference_pipe}. Like VQVAE2, our encoder first encodes the concatenated video input framewise into $32 \times 32$ and $64 \times 64$ dimensional top and bottom hierarchies, respectively. Before the quantization step at each of these hierarchies, our temporal modules are added that process the reduced video frames. This step allows the network to backpropagate with temporal connections between the frames. The further processing is then again done framewise using a standard VQVAE2 decoder. In practice, we observed that this temporal module plays an important role in generating temporally coherent outputs, as we show through ablations in Sec.~\ref{sec:ablation}. Our special autoencoder differs from standard autoencoders in the loss computation step. Instead of reconstructing the inputs, a six-channel video input -- the first three channels belonging to the source foreground and the last three channels belonging to the target pose and background -- FaceOff aims to generate a three-channel blended video output. Therefore, the loss computation is between a ground truth three-channel video and the three-channel video output.

\subsection{Self-supervised Training Approach}

\begin{figure}[t]
  \centering
  \includegraphics[width=1.\linewidth]{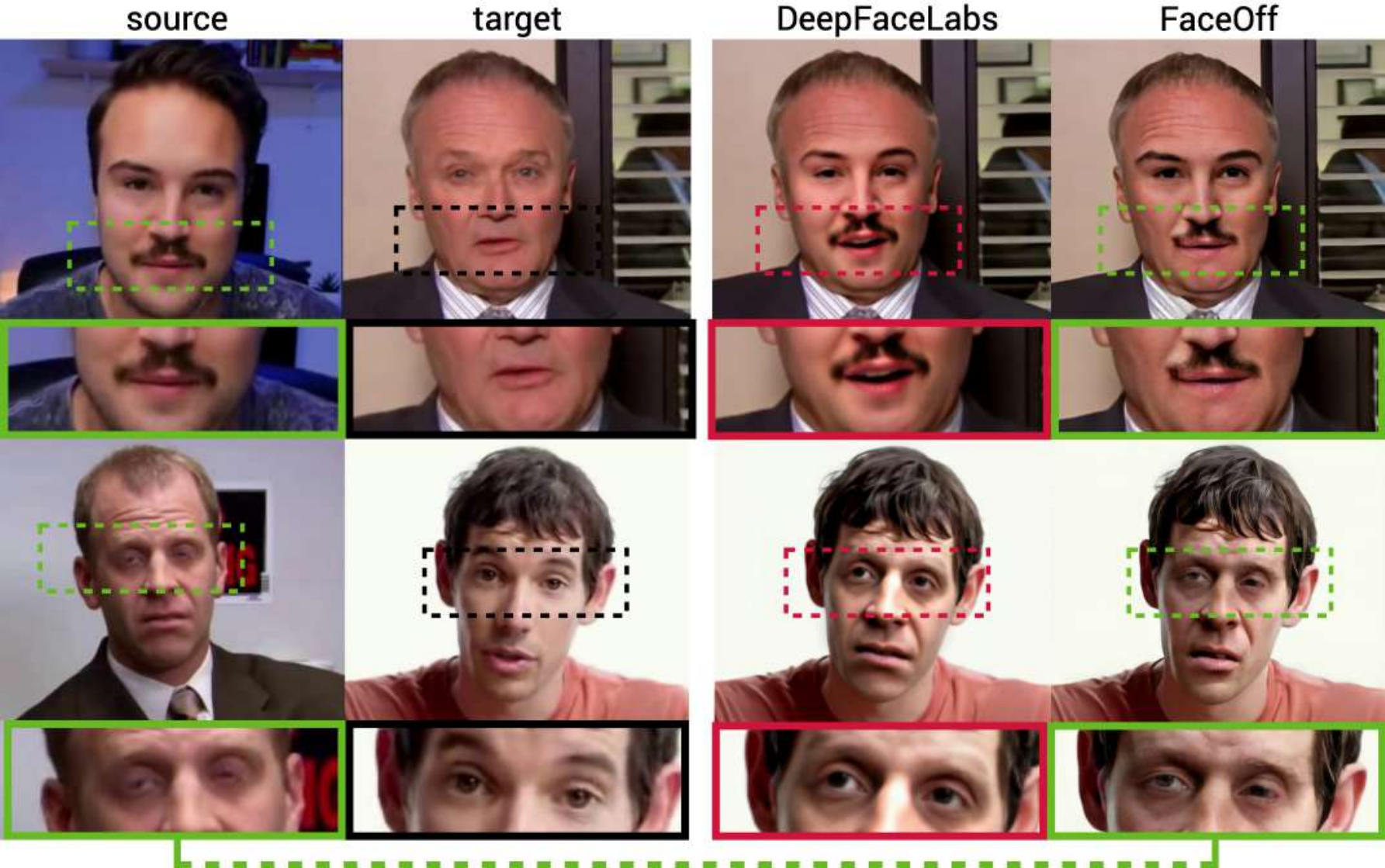}
  \caption{Existing face-swapping methods~\cite{deepfacelabs, motion-coseg, fsgan} use a generator-discriminator training strategy. This results in outputs with novel expressions as explained in Sec.~\ref{sec:self-supervised-explanation}. We show this phenomenon on DeepFaceLabs~\cite{deepfacelabs}. The expressions in the output (red boxes) do not match either of the inputs, source, or target. E.g., direction of eye gaze (second row) or overall laugh expression (first row). FaceOff successfully preserves the source expressions (green boxes).}
  \label{fig:expressions_mismatch}
\end{figure}

\label{sec:self-supervised-explanation}
Existing face-swapping approaches employ generators and discriminators to train their networks. These discriminators are classifiers that indicate a relationship between the generator's outputs and underlying data distribution, such as an identity or an expression distribution. In such a setup, the generators are encouraged to hallucinate some aspects of the outputs to match the discriminator's data distribution causing it to output novel identities or expressions. We show this phenomenon in Fig.~\ref{fig:expressions_mismatch}. A hard distance loss (e.g., Euclidean distance) indicating the exact output-ground truth relationship instead of a stochastic discriminator loss can be used to overcome this issue. In V2V face-swapping, retaining the exact source expressions is essential. Thus, we train our network using a distance loss by devising a self-supervised training scheme that forces the network to reconstruct a denoised version of a given input video. 

To understand the training scheme, we first look at the challenges we encounter when trying to blend two motions naively. First, there is a global and local pose difference between the faces in the source and target videos. We fix the global pose difference by aligning (rotating, translating, and scaling) the source poses according to the target poses using face landmarks, as shown in Fig.~\ref{fig:inference_pipe}. However, the local pose difference is not overcome this way, and we observe temporal incoherence across the frames. Next, we observe a difference in the foreground and background color (illumination, hue, saturation, and contrast). Thus, we train our network to solve these known issues by reproducing these errors during training. As illustrated in Fig.~\ref{fig:arch_main}, we train our model in the following manner:  (1) Take a video, say $s$. (2) From $s$, extract the face region, say $f$; and the background region, say $b$. (3) Introduce pseudo errors (rotation, color, scale, etc.) on $f$. (4) Construct the input $v$ by concatenating $f$ and $b$ channel-wise at every corresponding frame. (5) Train the network to construct $s$ from $v$. Although we train the network using the same identity in the self-supervised scheme, it can face-swap unseen identities directly at inference without any finetuning. 

\subsection{Reproducing Inference Errors at Training}

Given two talking-head videos, source and target, denoted by $S$ and $T$, respectively, our aim is to generate an output that preserves (1) the identity and the emotions from $S$ and (2) the pose and background from $T$. We assume the number of frames, denoted by $N$, in $S$ and $T$ are equal. Given two frames, $s_i \in S$ and $t_i \in T$ such that $i = 1 ... N$, we denote $f_{s_i} \in F_s$ and $b_{t_i} \in B_t$ as the foreground and background of $s_i$ and $t_i$, respectively. Given $F_s$ and $B_t$ as input, the network fixes the following issues:

First, the network encounters a local pose difference between $f_{s_i}$ and $b_{t_i}$. This pose difference can be fixed using an affine transformation function: $\delta(f_{s_i}, b_{t_i}) = m(rf_{s_i} + d) + m(rb_{t_i} + d)$ where $m$, $r$, and $d$ denote scaling, rotation, and translation. Face being a non-rigid body; the affine transformation only results in the two faces with a perfect match in the pose but a mismatch in shape. One can imagine trying to fit a square in a circle. One would need a non-linear function to first transform the square to a shape similar to the circle so that they fit. We denote this non-linear transformation as a learnable function $\omega(f_{s_i}, b_{t_i})$. Being non-linear, a network can perform such transformations on the input frames as long as both faces fit. These transformations can be constrained using a distance loss to encourage spatially-consistent transformations that generate a meaningful frame. However, these spatially-consistent transformations may be temporally-incoherent across the video. This would result in a video with a face that wobbles, as shown in Sec.~\ref{sec:ablation}. Thus, we constrain the transformations as $\omega(f_{s_i}, b_{t_i}, f_{s_k}, b_{t_k})$ where $k = 1..N$ such that $k \ne i$. Here, the transformation on the current frame is constrained by the transformations on all the other frames in the video. This is enabled by the temporal module, as explained in Sec.~\ref{sec:merging_video_using}. 
Lastly, the network encounters a difference in color between $f_{s_i}$ and $b_{t_i}$ that is fixed as $c(f_{s_i}, b_{t_i})$.

\begin{figure*}[t]
  \centering
  \includegraphics[width=1.0\linewidth]{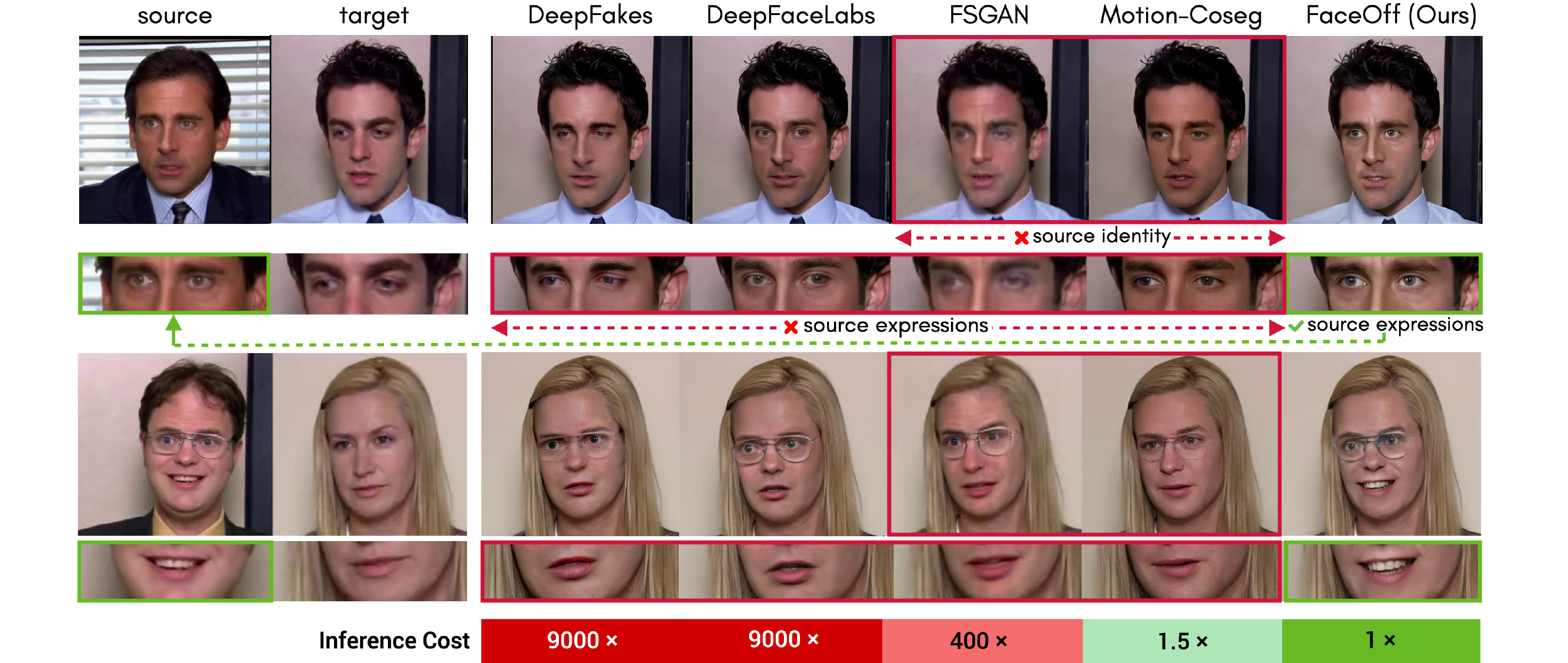}
  \caption{``Inference Cost" denotes the time taken for a single face-swap. 
  FSGAN, with $400\times$ FaceOff's inference cost, fails to swap the identities fully. 
  DeepFakes and DeepFaceLabs swap the identities successfully but are $9000\times$ less efficient than FaceOff. FaceOff perfectly swaps source identity and expressions. None of the other methods can swap source expressions. }
  \label{fig:im2im}
\end{figure*}

\begingroup
\renewcommand{\arraystretch}{1.1} 
\begin{table*}[t]
  \centering
  \adjustbox{max width=0.93\textwidth}{
  \begin{tabular}{l|ccccc|ccc}
  \hline
    & \multicolumn{5}{ c|  }{\textbf{Quantitative Evaluation}} & \multicolumn{3}{c}{\textbf{Human Evaluation}}\\
    \cline{2-9}
    \textbf{Method} & \textbf{SPIDis} $\downarrow$ & \textbf{LMD} $\downarrow$ & \textbf{TL-ID} $\uparrow$ & \textbf{TG-ID} $\uparrow$ & \textbf{FVD} $\downarrow$  &  \textbf{Identity} $\uparrow$ & \textbf{Exps.} $\uparrow$ & \textbf{Ntrl.} $\uparrow$ \\
    \hline
    Motion-coseg~\cite{motion-coseg} & $0.48$ & $0.59$ & $0.872$ & $0.893$ & $293.652$ & $6.82$ & $5.81$ & $7.44$ \\
    FSGAN~\cite{fsgan} & $0.49$ & $0.57$ & $0.914$ & $\mathbf{0.923}$ & $\mathbf{242.691}$ & $7.84$ & $6.83$ & $\mathbf{8.31}$ \\
    FaceOff ( Ours ) & $\mathbf{0.38}$ & $\mathbf{0.41}$ & $\mathbf{0.925}$ & $0.915$ & $255.980$ & $\mathbf{9.64}$ & $\mathbf{9.86}$ & $8.18$ \\
    \hline
  \end{tabular}
  }
    \caption{Quantitative metrics on V2VFaceSwap dataset. DeepFakes and DeepFaceLabs take up to $24$ hours for best inference on a single face-swap~\cite{deepfacelabs}; thus, we do not compare with them. The metrics used for comparisons are explained in Sec.~\ref{sec:experiments}. For fair comparisons, FSGAN scores are reported without any inference time optimization. Although FSGAN has a slightly better FVD and Naturalness (Ntrl.) score, it fails to swap the identity fully, as can be clearly seen from SPIDis, LMD, and Identity metric. Moreover, the difference in the FVD of FSGAN and FaceOff is not statistically significant perceptually~\cite{fvd}.}
     \label{tab:metrics}
\end{table*}
\endgroup

As shown in Fig.~\ref{fig:arch_main}, at the time of training $S=T$. For each frame $s_i \in S$, we first extract the foreground, $f_{s_i} \in F_s$ (acting as the source), and the background, $b_{t_i} \in B_t$ (acting as the target) from $s_i$. Next, we apply random rotation, translation, scaling, color, and distortion (Barrel, Mustache) errors on $f_{s_i}$. The training setting is then formulated as: 


\begin{gather}
    \Phi: \Omega(\delta, \omega, c)\\
    J = \frac{1}{N}\sum_{i = 1}^{N} [ s_i - \Phi(f_{s_i}, b_{t_i}, f_{s_k}, b_{t_k})] + P(F_s, B_t)
\end{gather}

where $\Omega$ is a learnable function, $J$ is the overall cost of the network to be minimized, and $P$ is a perceptual metric (LPIPS~\cite{lpips} in our case), and $k = 1\dots N$ such that $k \neq i$.

\section{Experiments and Results}
\label{sec:experiments}

\begin{figure*}[t]
  \centering
  \includegraphics[width=0.9\linewidth]{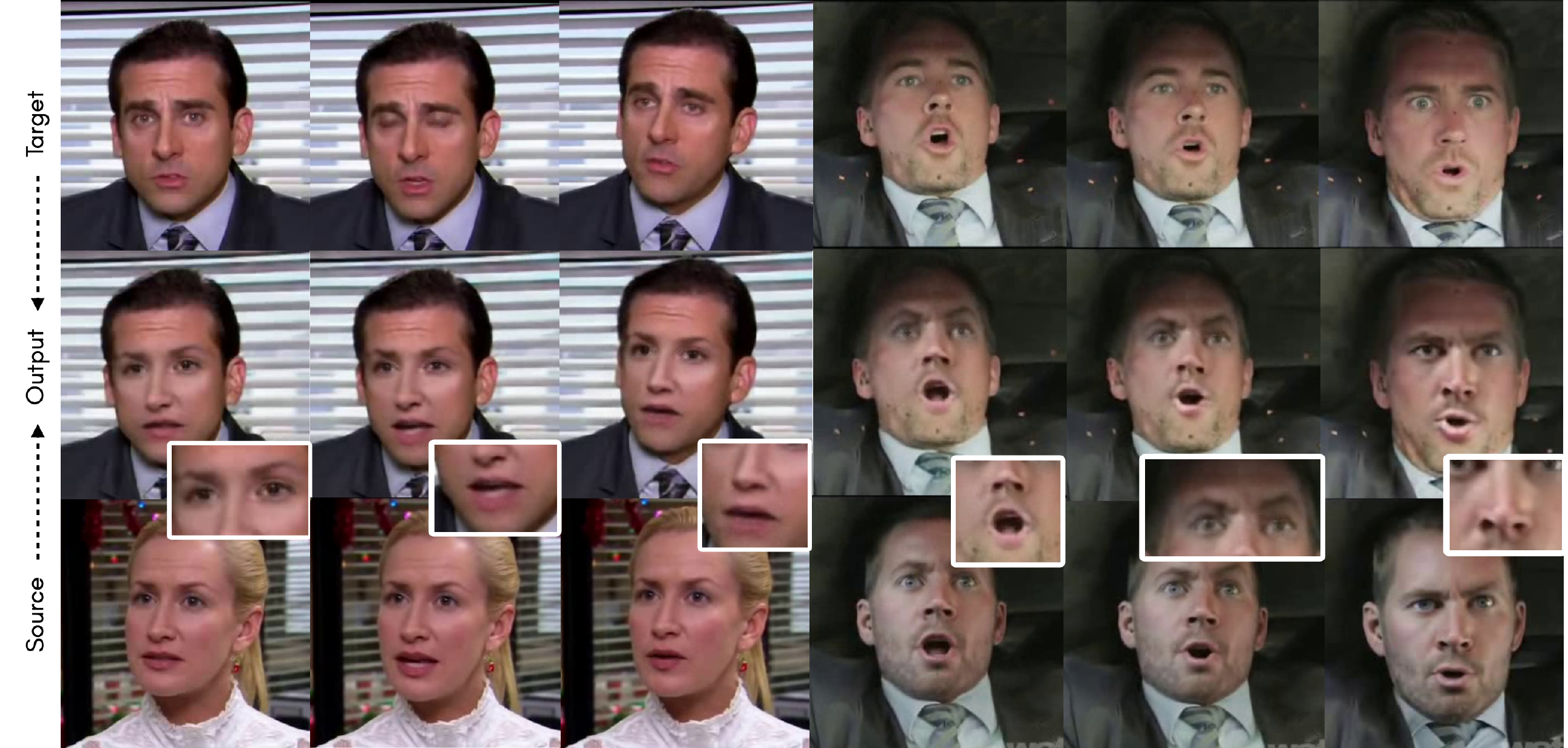}
  \caption{Qualitative results of FaceOff. Note that there is a significant difference in the source and target expressions in all the cases. FaceOff swaps the source expressions (mouth, eyes, etc.) and identity; and retains the target pose and background.} 
  \label{fig:main_results}
\end{figure*}

In this section, we try to answer the following questions: (1) How well can we preserve the source identity compared to the alternate approaches? (2) How well do we preserve the expressions of the input videos? (3) How efficient is FaceOff when compared to other techniques? 

We compare FaceOff against different tasks: ``face-swapping", ``face reenactment", and ``face editing". Please note that none of these methods can fully solve the task of V2V face-swapping that we aim to solve. Specifically, V2V face-swapping aims to (1) swap source identity and expressions and (2) retain the target pose and background.

\textbf{Quantitative Metrics: } 
\textbf{(1)} \textbf{S}ource-\textbf{P}rediction \textbf{I}dentity \textbf{Dis}tance \textbf{(SPIDis)}: computes the difference in identity between face images. It is computed as the Euclidean distance between the face embeddings generated using dlib's face detection module.
\textbf{(2)} \textbf{F}réchet \textbf{V}ideo \textbf{D}istance \textbf{(FVD)}, as proposed in \cite{fvd}, computes the temporal coherence in the generated video output. \textbf{(3)} \textbf{L}and\textbf{m}ark \textbf{D}istance \textbf{(LMD)}: evaluates the overall face structure and expressions of the source and swapped output. To compute LMD, the source and the swapped face landmarks are normalized: faces are first centered and then rotated about the x-axis so that the centroid and angle between the eye coordinates align a mean image. Next, the faces are scaled to the mean image. Euclidean distance between the normalized swapped and source video landmarks gives the LMD. We compute LMD between the source and the output face expressions (excluding the landmarks of the face permiter). 
\textbf{(4)} \textbf{T}emporally \textbf{L}ocally \textbf{(TL-ID)} and \textbf{T}emporally \textbf{G}lobally \textbf{(TG-ID)} \textbf{Id}entity
Preservation: proposed in \cite{stit}. They evaluate a video's identity consistency at a local and global level. For both metrics, a score of 1 would indicate that the method successfully maintains the identity consistency of the original video.

\textbf{Qualitative Metrics: }A mean absolute opinion score on a scale of $1-10$ is reported for \textbf{(1)} \textbf{Identity}: How similar is the swapped-output identity with the source identity? \textbf{(2)} Expressions \textbf{(Exps.)}: How similar is the swapped-output expression with the source expression?, and \textbf{(3)} Naturalness \textbf{(Ntrl.)}: Is the generated output natural?

\textbf{Experimental Dataset}: We benchmark the V2VFaceSwap dataset made of unconstrained YouTube videos with many unseen identities, backgrounds, and lighting conditions. The supplementary paper reports further details about the dataset and evaluation setup.


\subsection{Face-Swapping Results}

Fig.~\ref{fig:im2im} and Table~\ref{tab:metrics} present a qualitative and quantitative comparison, respectively, between the existing methods and FaceOff. Fig.~\ref{fig:main_results} demonstrates FaceOff's face-swapping results on videos. 
As shown in Fig.~\ref{fig:im2im}, FaceOff successfully swaps the identity and expressions of the source face video. Existing methods cannot swap the source expressions, which shows that FaceOff solves a unique challenge of V2V face-swapping. An interesting finding of our experiments is that the existing methods do not preserve any input expressions -- source or target -- at the output and generate novel expressions, e.g., novel gaze direction or mouth movements. This phenomenon is also demonstrated in Fig.~\ref{fig:expressions_mismatch}. FSGAN and Motion-Coseg fail to swap the identity entirely. This is further corroborated through quantitative metrics in Table~\ref{tab:metrics}. FaceOff shows an improvement of $\sim 22\%$ and $\sim 28\%$ on SPIDis and LMD over FSGAN, indicating FaceOff's superiority.

\begin{figure}[t]
  \centering
  \includegraphics[width=0.93\linewidth]{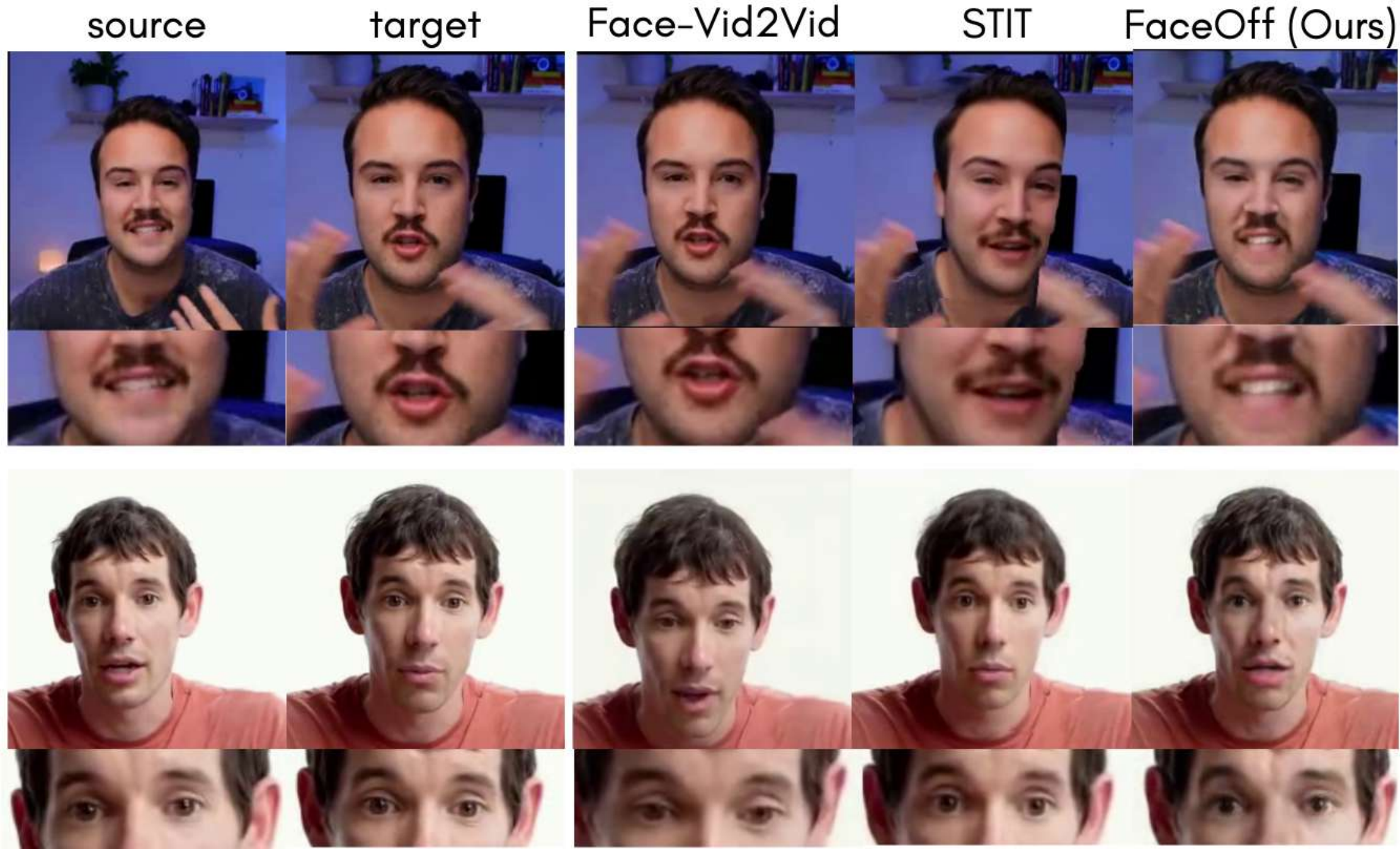}
  \caption{ Qualitative demonstration of Face Manipulation. As can be seen, none of the methods, except FaceOff, preserve the source expressions or pose information perfectly.}
  \label{fig:facemani}
\end{figure}

FSGAN achieves a slightly better FVD and is voted more natural in human evaluation. This is expected as FSGAN does not change the target identity much and retains the original target video making it more natural to observe. FaceOff swaps identity near-perfectly. Moreover, existing methods only have a single target motion to follow. FaceOff tackles an additional challenge of motion-to-motion swapping that needs source-target pose alignment at every frame. This requires FaceOff to generate a novel motion such that the identity, expressions, and pose in the motion look natural and match the inputs. Despite this challenge, the difference between FSGAN and FaceOff's FVD is not perceptually significant \cite{fvd}.
DeepFaceLabs and DeepFakes swap identity well but are $9000\times$ more computationally expensive than FaceOff, making FaceOff much more scalable and applicable in the real world.

\subsection{Target Face Manipulation Results}
\label{sec:targetfacemanipulation}

Given that the source and target have the same identity, the problem reduces to the following - transfer expressions from a source video to a target video. This is fundamentally the setting of ``face reenactment." One could also modify the expression of the target by identifying and quantifying the source expressions and using a ``face-editing" network to edit the target expressions. Fig.~\ref{fig:facemani} presents a qualitative comparison between FaceOff, ``face reenactment" (Face-Vid2Vid) and ``face-editing" (STIT).

\textbf{Face Reenactment}:  
We compare against Face-Vid2Vid~\cite{face-vid2vid}, a SOTA face reenactment network that reenacts the pose and expression of a target image using a source (driving) video. As shown in Fig.~\ref{fig:facemani}, FaceOff preserves the source's micro-expression, such as exact mouth opening and eye-frown. FaceOff relies on a deterministic distance loss, so it can retain the precise input expressions in the output. Moreover, FaceOff retains the temporal target pose and background, whereas Face-Vid2Vid modifies a static frame.

\textbf{Face Editing: } 
Using a powerful neural network, one can introduce the desired expressions in a video by performing edits. 
We compare our method against STIT~\cite{stit}. STIT modifies the expressions of a face video based on an input label. We observe the source expression and manually try out various intensities of the "smile" emotion ranging from negative to positive direction.
As seen in Fig.~\ref{fig:facemani}, although STIT can change the overall expression, it needs a manual hit-and-trial to pinpoint the exact expression. It also lacks personalized expression (amount of mouth opening, subtle brow changes). Also, each and every expression cannot be defined using a single label, and introducing variations in emotion along the temporal dimension is hard. With our proposed method, one can incorporate any emotion in the video (as long as we have access to a source video). 

\section{Ablation Study}
\label{sec:ablation}

\begin{figure}[t]
  \centering
  \includegraphics[width=0.9\linewidth]{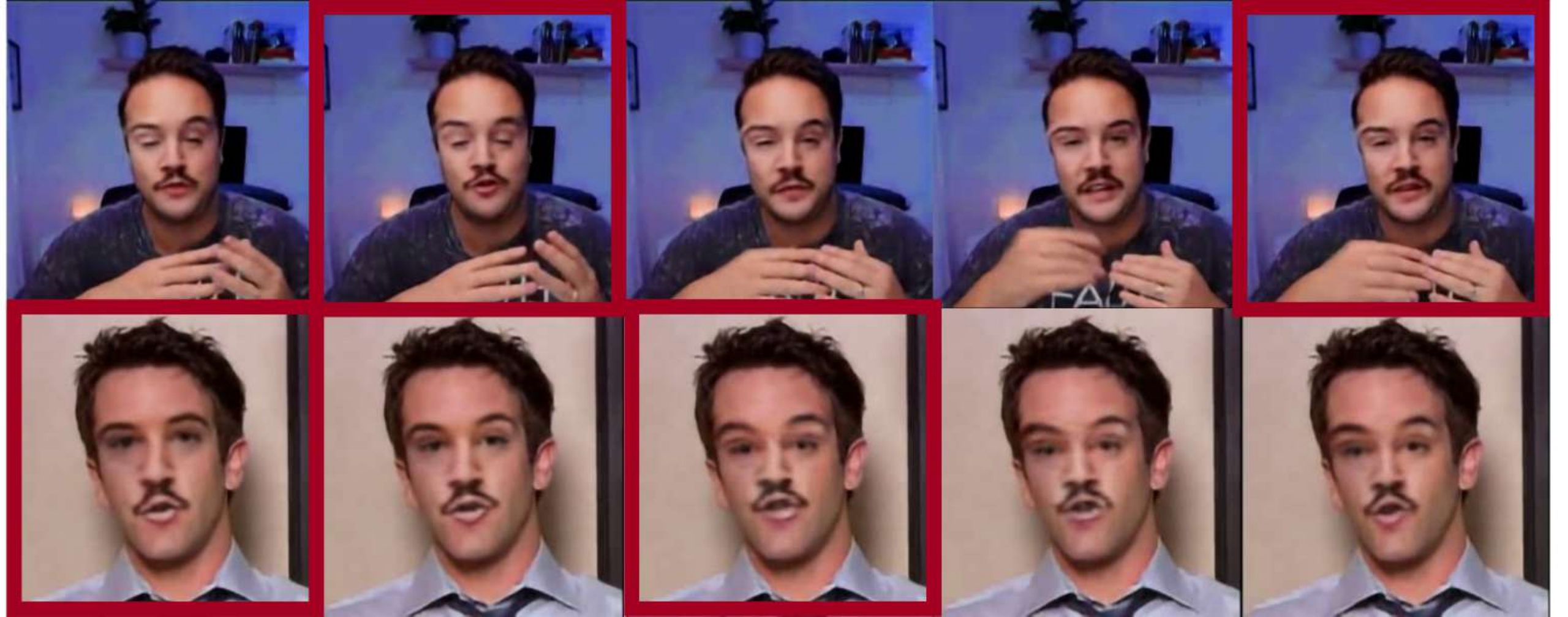}
  \caption{\small FaceOff without Temporal Module. As we jump from one frame to another (red boxes), we can observe a "wobble effect": significant change in the facial structure (elongated and then squeezed). This occurs as the model does not have an understanding of the neighboring frames while generating the current frame.}
  \label{fig:ablation}
\end{figure}

We investigate the contribution of different modules and errors in achieving FaceOff. 
Fig.~\ref{fig:ablation} demonstrates the performance of FaceOff without the proposed temporal module. As shown, although at a frame level, the output is spatially-coherent, as we look across the frames, we can notice the temporal incoherence. The face seems to `wobble' across the frames - squishing up and down. In fact, without the temporal module, the network does not understand an overall face structure and generates unnatural frames (marked in red). Jumping from one red box to another, we can see that the face structure has completely changed. This suggests that constraining the network by the neighboring frames using the temporal module enables the network to learn a global shape fitting problem, consequently generating a temporally coherent output. 

Table~\ref{tab:ablation} presents the quantitative contribution of the temporal module and each of the errors used for self-supervised training. The metrics indicate that each of them contributes significantly to achieving FaceOff.

\begin{table}[t]
    \centering
    \adjustbox{max width=0.8\linewidth}{
    \begin{tabular}{l|ccc}
        \toprule
        Component & SPIDis $\downarrow$  & LMD $\downarrow$ & FVD $\downarrow$  \\
        \midrule
        FaceOff & $\mathbf{0.38}$ & $\mathbf{0.41}$ & $\mathbf{255.980}$ \\
        \midrule
        w/o Temporal.  & 0.71 & 0.49 & 350.60 \\ 
        w/o Rotation & 0.65 & 0.44 & 292.76 \\
        w/o Color & 0.74 & 0.42 & 303.35 \\
        w/o Translation & 0.58 & 0.47 & 271.82 \\
        w/o Distortion & 0.55 & 0.45 & 285.54 \\
        \bottomrule
    \end{tabular}
    }
    \caption{\small 
    We remove different components and errors and evaluate their contributions in achieving FaceOff.  }
    \label{tab:ablation}
\end{table}

\section{Conclusion}
We introduce ``video-to-video (V2V) face-swapping", a novel task of face-swapping. Unlike face-swapping, which aims to swap an identity from a source face video (or an image) to a target face video, V2V face-swapping aims to swap the source expressions along with the identity.
To tackle this, we propose FaceOff, a self-supervised temporal autoencoding network that takes two face videos as input and produces a single coherent blended output. As shown in the experimental section, FaceOff swaps the source identity much better than the existing approaches while also being $400\times$ computationally efficient. It also swaps the exact source identity that none of the methods can do. V2V face-swapping has many applications; a significant application can be automating the task of replacing the double's face with the actor's identity and expressions in movies. We believe our work adds a whole new dimension to movie editing that can potentially save months of tedious manual effort and millions of dollars.

{\small
\bibliographystyle{ieee_fullname}
\bibliography{egbib}
}

\end{document}